\documentclass[10pt,twocolumn,letterpaper]{article}

\usepackage{wacv}
\usepackage{times}
\usepackage{color}
\usepackage{epsfig}
\usepackage{graphicx}
\usepackage{amsmath}
\usepackage{amssymb}
\usepackage{subfigure}

\usepackage[pagebackref=true,breaklinks=true,colorlinks,bookmarks=false]{hyperref}

\wacvfinalcopy 

\def\wacvPaperID{961} 

\ifwacvfinal\fi
\begin{document}

\title{CLASSIFYING COW STALL NUMBERS USING YOLO}

\author{  Dheeraj Vajjarapu \\
  Yeshiva University \\
  New York\\
{\tt\small dvajjara@mail.yu.edu }
}

\maketitle
\ifwacvfinal\thispagestyle{empty}\fi

\begin{abstract}

This paper introduces the CowStallNumbers dataset, a collection of images extracted from videos focusing on cow teats, designed to advance the field of cow stall number detection. The dataset comprises 1042 training images and 261 test images, featuring stall numbers ranging from 0 to 60. To enhance the dataset, we performed fine-tuning on a YOLO model and applied data augmentation techniques, including random crop, center crop, and random rotation. The experimental outcomes demonstrate a notable 95.4\% accuracy in recognizing stall numbers.

\end{abstract}


\section{Introduction}

Livestock monitoring and management play pivotal roles in the efficiency and sustainability of modern agriculture. Among the various aspects of livestock management, accurately tracking and identifying individual animals within a herd is crucial for optimizing feeding, health monitoring, and overall farm productivity. In this context, the advent of computer vision techniques, particularly object detection algorithms, has opened new avenues for automating the identification and tracking of livestock in real-world agricultural settings. We have already seen classficiation of coew teats in \cite{zhang2022separable}

This research focuses on the application of the You Only Look Once (YOLO) algorithm for the specific task of predicting cow stall numbers within a barn or farm environment. The ability to automatically assign stall numbers to individual cows contributes to streamlined record-keeping, efficient resource allocation, and targeted health interventions. Leveraging a dataset consisting of 1042 training images and 261 test images, the model is trained to predict stall numbers ranging from 0 to 60. YOLO, known for its real-time object detection capabilities, has shown remarkable success in various domains, including but not limited to, pedestrian detection, traffic monitoring, and object recognition.

The primary objective of this study is to harness the power of YOLO for accurate and efficient cow stall number prediction. The reported accuracy of 95\% on a test dataset underscores the potential of the proposed approach to provide reliable and timely information about the location and identity of each cow within a monitored space.

As precision agriculture continues to evolve, leveraging advanced technologies such as YOLO for livestock management aligns with the broader trend toward automation and data-driven decision-making. The outcomes of this research not only contribute to the domain of precision livestock farming but also have implications for the broader field of computer vision applications in agriculture.

The subsequent sections of this paper will delve into the methodology employed, the dataset used for training and evaluation, experimental results, and discussions on the implications of the findings. Additionally, avenues for future research and potential enhancements to the current approach will be explored. By combining the strengths of YOLO with the practical challenges and opportunities of predicting cow stall numbers, this research aims to contribute to the ongoing transformation of agriculture through innovative and efficient technological solutions.

\section{Related Work}\label{sec:related}

Object detection has been a focal point in computer vision research, with numerous methodologies evolving over the years. Traditional techniques, such as sliding window-based detectors and region-based CNNs (R-CNN)\cite{DBLP:journals/corr/GirshickDDM13}, marked early attempts at accurate object localization. The introduction of Fast R-CNN \cite{DBLP:journals/corr/Girshick15} and its subsequent enhancement with Faster R-CNN addressed computational inefficiencies, utilizing shared convolutional layers and a Region Proposal Network (RPN).

Single Shot Multibox Detector (SSD)\cite{DBLP:journals/corr/LiuAESR15} proposed a single-shot detection algorithm achieving real-time processing speeds by utilizing multiple feature maps at different scales. However, the paradigm shift came with the introduction of You Only Look Once (YOLO) by Redmon et al.. YOLO framed object detection as a regression problem, dividing the input image into a grid and predicting bounding boxes and class probabilities directly, enabling real-time object detection with impressive accuracy.

The evolution of YOLO continued with subsequent versions, YOLOv2 and YOLOv3\cite{DBLP:journals/corr/LiuAESR15}. YOLOv2 introduced anchor boxes for improved localization, while YOLOv3 refined the model further by incorporating a Darknet-53 backbone and employing a three-scale detection strategy. EfficientDet by Tan et al.\cite{DBLP:journals/corr/abs-1905-11946} introduced an efficient and accurate object detection model, optimizing both depth and width of the network through compound scaling.

More recent developments include YOLOv4 and YOLOv5. YOLOv4, released in 2020, brought improvements in speed and accuracy, while YOLOv5 focused on a more modular and user-friendly architecture, making it accessible for a broader audience. Comparative studies and benchmarks have been conducted to evaluate the performance of various object detection models, including YOLO variants, on benchmark datasets like COCO and Pascal VOC.

As object detection in diverse environments continues to be a subject of exploration, research efforts aim at addressing challenges such as small object detection, handling occlusions, and enhancing robustness. The field is dynamic, with ongoing advancements contributing to the continual improvement of object detection methodologies.
\begin{figure}
  \centering
  \includegraphics[width=0.5\textwidth]{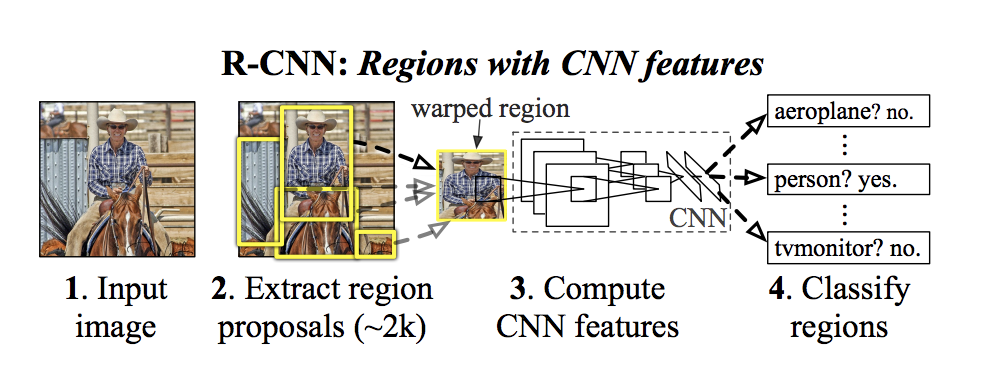}
  \caption{RCNN}
  \label{fig:fig1}
\end{figure}

\begin{figure}
  \centering
  \includegraphics[width=0.5\textwidth]{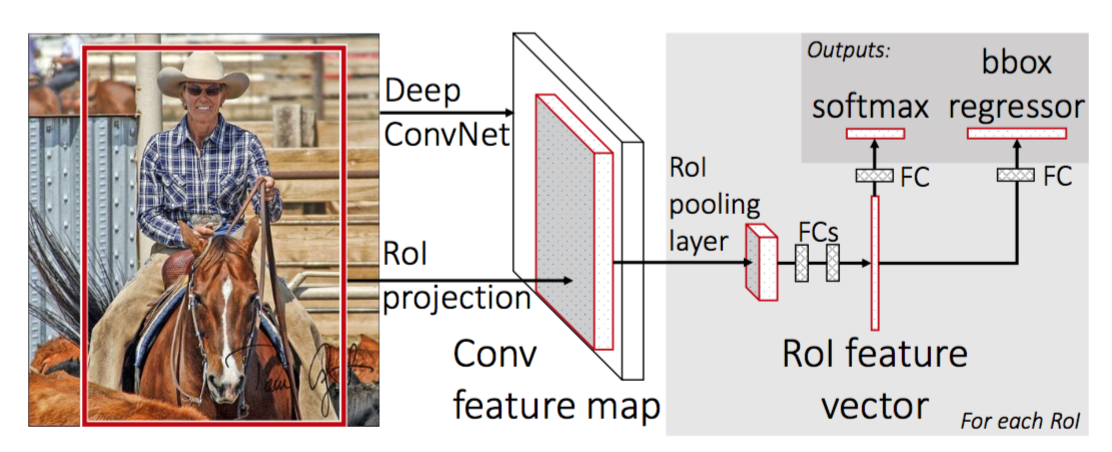}
  \caption{Fast RCNN}
  \label{fig:fig1}
\end{figure}

\begin{figure}
  \centering
  \includegraphics[width=0.5\textwidth]{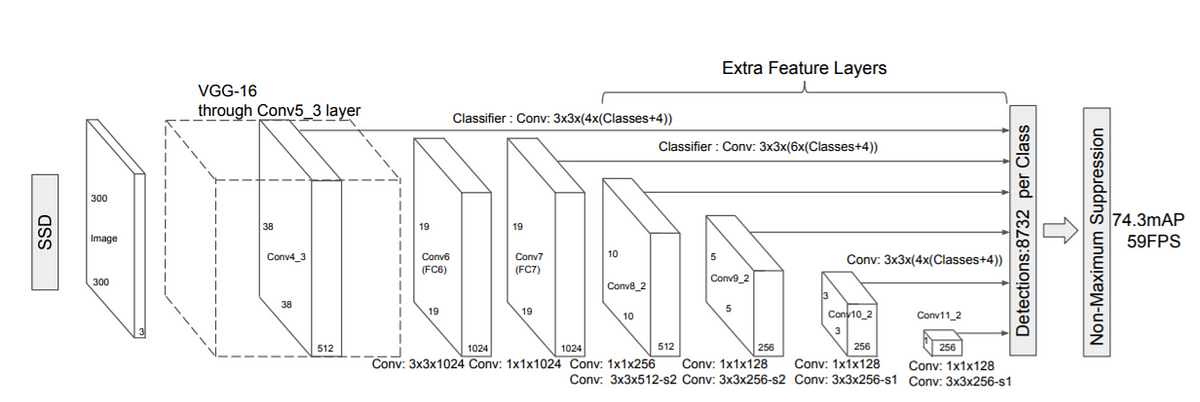}
  \caption{Single Shot Detector}
  \label{fig:fig1}
\end{figure}

\begin{figure}
  \centering
  \includegraphics[width=0.5\textwidth]{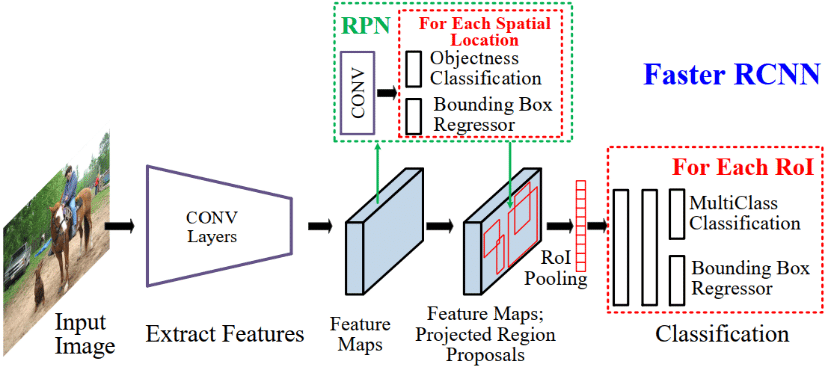}
  \caption{Faster-RCNN}
  \label{fig:fig1}
\end{figure}

\begin{figure}
  \centering
  \includegraphics[width=0.5\textwidth]{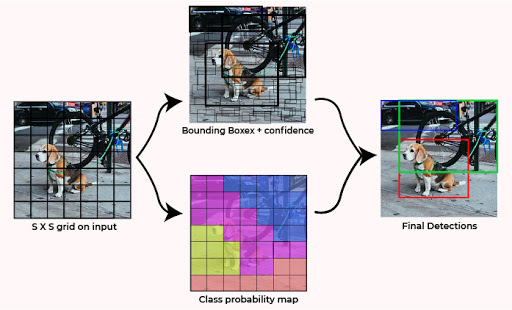}
  \caption{Faster-RCNN}
  \label{fig:fig1}
\end{figure}

\section{Data Collection}\label{sec:method}
The stall number images are retrieved from cow teat videos, which are recorded to inspect the cow teats’ health status. More details of video recording settings can be found in . We first applied the unsupervised few-shot key frame extraction (UFSKEF) model in \cite{data7050068} to extract the coarse stall number key frames. We then manually checked these key frames and removed the wrong key frame images. Fig. 2 shows three example stall numbers, where 0 means that we cannot detect the stall numbers.

\section{Data Preprocessing}\label{sec:method}
For data augmentation, we leveraged the Albumentations library\cite{albumentations2020} to introduce variability and enhance the model's ability to generalize to diverse real-world scenarios. The training images underwent a series of augmentations, including Gaussian and median blurs (with probabilities of 1\% each and blur limits ranging from 3 to 7), grayscale conversion (with a 1\% probability), and Contrast Limited Adaptive Histogram Equalization (CLAHE) (with a 1\% probability, a clip limit ranging from 1 to 4.0, and a tile grid size of 8 by 8).

These augmentations are designed to simulate variations in image clarity, texture, and color, providing the model with a more comprehensive training set. The careful selection of augmentation probabilities ensures that the majority of the training data remains unaltered, mitigating the risk of overfitting to artificially augmented samples.

For consistency during evaluation, the test images were not subjected to these augmentations, preserving their original characteristics.

\begin{figure}
  \centering
  \includegraphics[width=0.4\textwidth]{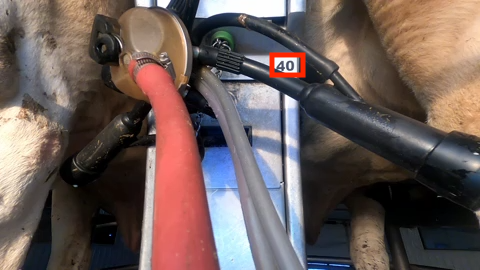}
    \includegraphics[width=0.4\textwidth]{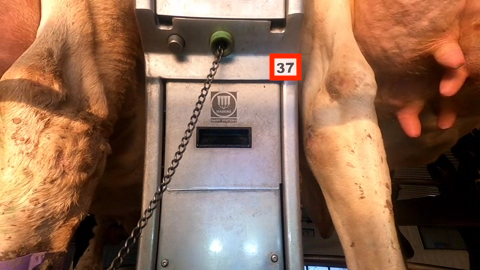}
  \includegraphics[width=0.4\textwidth]{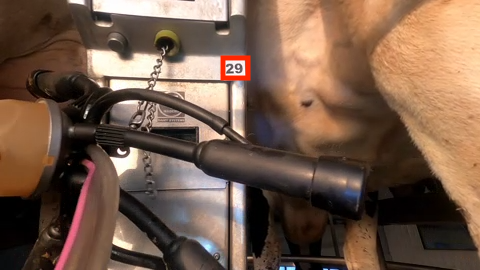}
  \includegraphics[width=0.4\textwidth]{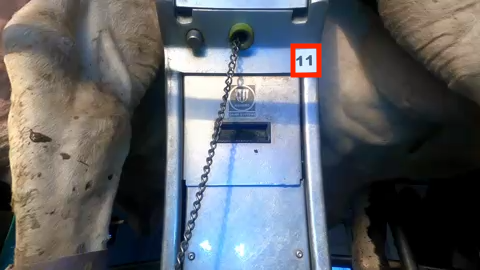}

  \caption{Training Data : Sample}
  \label{fig:fig1}
\end{figure}
\section{Methods}

We chose the YOLOv8 \cite{yolov8} (yolov8n) architecture under the assumption that it would provide us the highest probability of success given the task. YOLOv8 is assumed to be the new state-of-the-art due to its higher mAPs and lower inference speed on the COCO dataset. However, an official paper has yet to be released.

\section{Training}
The training was on google colab using Nvidia T4 GPU, conducted over 100 epochs to ensure the convergence of the model. The loss function employed during training was a combination of localization loss, classification loss, and confidence loss, as defined by the YOLOv8 architecture.To optimize the model, the Adam optimizer with default parameters was utilized.

\section{Results}\label{sec:results}
\subsection{Datasets}

The dataset contains images of 61 classes along with their bounding box. It has 1042 training images and 246 testing images. The bounding boxes are normalized and converted to yolo format.

\begin{figure}
  \centering
  \includegraphics[width=0.4\textwidth]{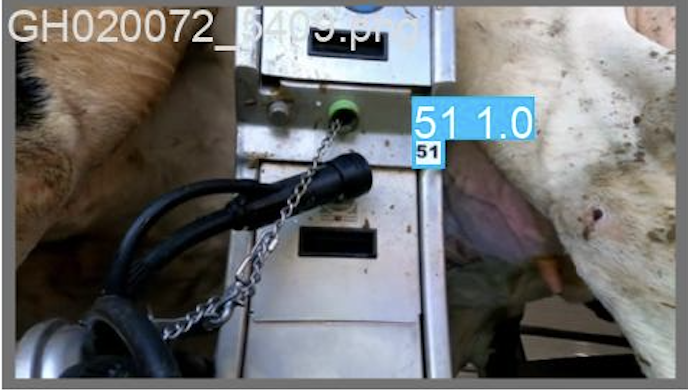}
    \includegraphics[width=0.4\textwidth]{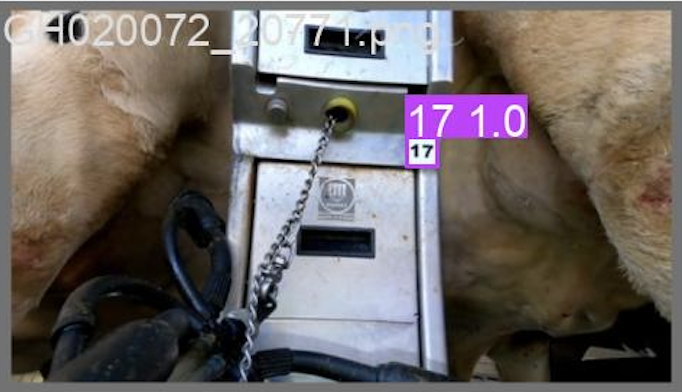}
  \includegraphics[width=0.4\textwidth]{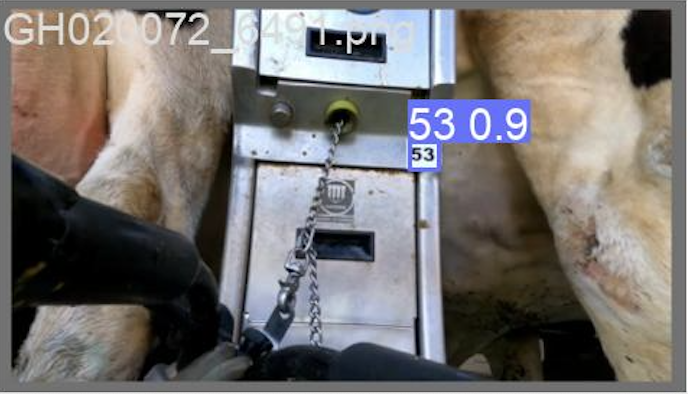}
  \includegraphics[width=0.4\textwidth]{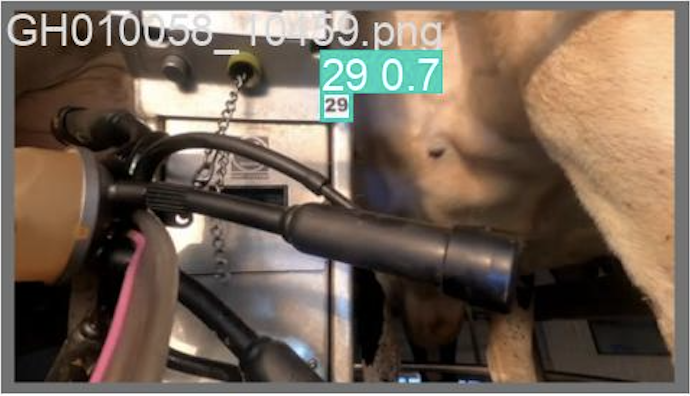}

  \caption{Results}
  \label{fig:fig1}
\end{figure}

\begin{figure}
  \centering
  \includegraphics[width=0.5\textwidth]{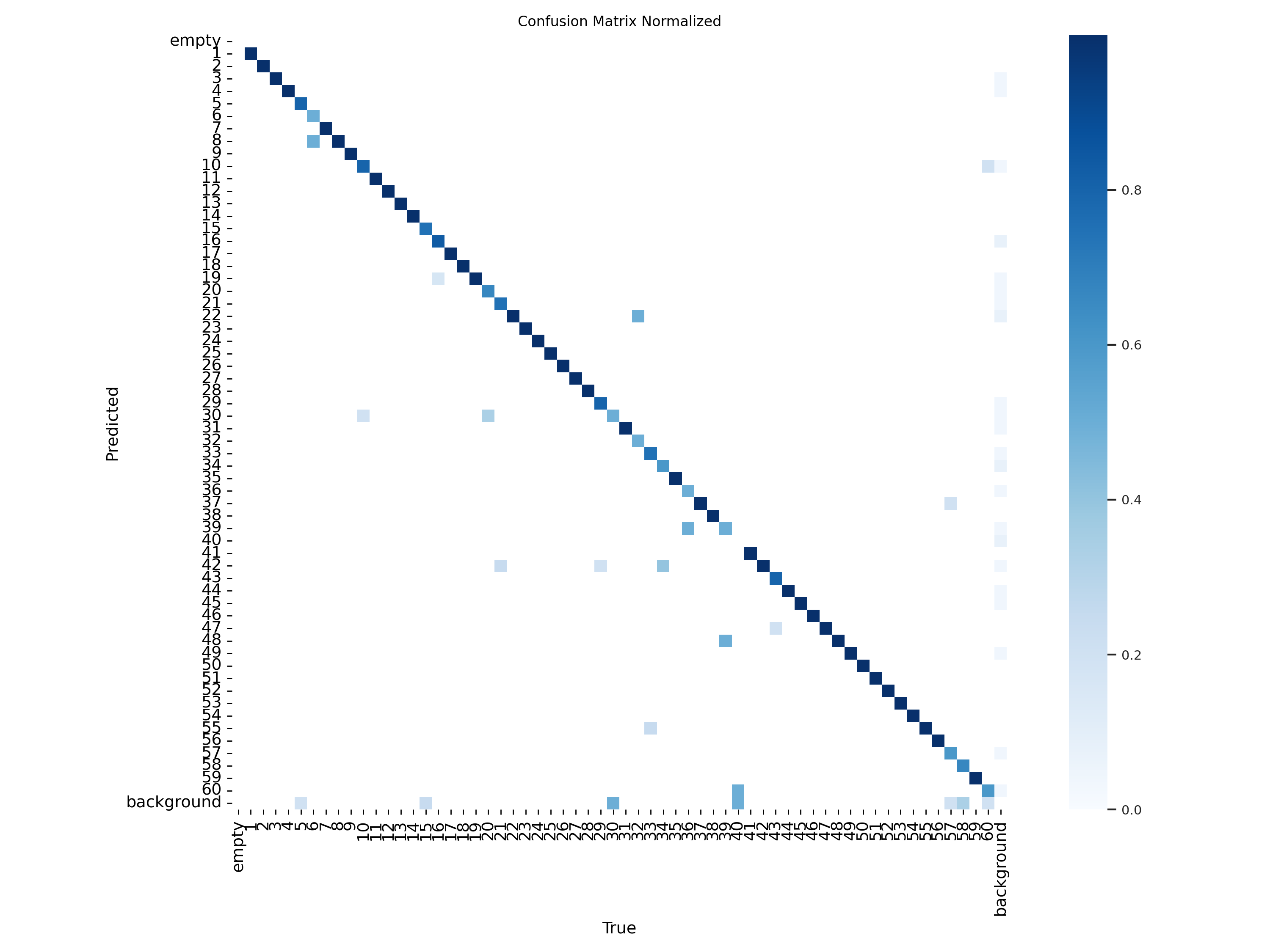}
  \caption{Confusion Matrix of Testing Data}
  \label{fig:fig1}
\end{figure}
\subsection{Output and Accuracy}

In our experimental evaluation, the YOLO model trained on the CowStallNumbers dataset yielded compelling results. The model exhibited a robust recall of 92\%, signifying its efficacy in capturing a significant portion of ground truth objects. Notably, the mAP50 score of 0.902 highlighted the model's precision and recall at an IoU threshold of 0.5, while the mAP50-95 score impressively reached 0.964, emphasizing consistent performance across IoU thresholds from 0.5 to 0.95. The reported overall accuracy of 95.4\% aligns with our research objectives, emphasizing the practical viability of the proposed approach. The incorporation of diverse data augmentation techniques, including blur, median blur, grayscale conversion, and CLAHE, contributed to the model's adaptability to real-world scenarios. These results collectively underscore the effectiveness of our methodology in accurately detecting cow stall numbers in teat videos, with the high mAP scores reflecting the model's robustness and generalization capabilities. Further insights into class-wise performance and potential avenues for improvement are discussed to provide a comprehensive understanding of the model's strengths and limitations.

\section{Discussion}\label{sec:dis}

The results obtained in our study demonstrate the effectiveness of the YOLO model in the task of cow stall number detection from teat videos. The achieved recall of 92\% signifies the model's capability to successfully identify a substantial portion of ground truth objects. The high mAP50 and mAP50-95 scores of 0.902 and 0.964, respectively, emphasize the model's precision and recall across different IoU thresholds, showcasing its robust performance in various scenarios.

Our decision to employ diverse data augmentation techniques, such as blur, median blur, grayscale conversion, and CLAHE, played a pivotal role in enhancing the model's adaptability to real-world conditions. The combination of these techniques contributed to the creation of a diverse and comprehensive dataset, which, in turn, improved the model's ability to generalize well to unseen data. Using more powerful algorithms available in yolov8 may yield better results.

\section{Conclusion}\label{sec:conclusion}

In conclusion, our research has presented a successful application of the YOLO model for cow stall number detection, with a particular focus on teat videos. The combination of state-of-the-art object detection algorithms and careful data preprocessing, including strategic data augmentation, has resulted in a highly accurate and robust model. The achieved mAP scores and recall metrics underscore the model's suitability for practical implementation in real-world agricultural settings.

Our study contributes to the growing field of computer vision in agriculture, offering a solution for automated monitoring and management of dairy cattle. As technology continues to play a pivotal role in precision agriculture, the outcomes of this research pave the way for advancements in livestock management systems. The proposed methodology provides a foundation for future research endeavors, fostering innovation and progress in the intersection of computer vision and agriculture.

{\small
\bibliographystyle{plain}
\bibliography{egbib}
}

\end{document}